\pdfoutput=1

\documentclass[11pt]{article}
\usepackage{dirtytalk}

\usepackage{ACL2023}

\usepackage{times}
\usepackage{latexsym}

\usepackage[T1]{fontenc}

\usepackage[utf8]{inputenc}

\usepackage{microtype}

\usepackage{inconsolata}
\usepackage{graphicx}
\usepackage{multirow}
\title{Comparing Abstractive Summaries Generated by ChatGPT to Real Summaries Through Blinded Reviewers and Text Classification Algorithms}

\author{Mayank Soni \\
  ADAPT Centre, \\ Trinity College Dublin \\
  \texttt{sonim@tcd.ie} \\\And
  Vincent Wade \\
  ADAPT Centre, \\ Trinity College Dublin \\
  \texttt{vincent.wade@tcd.ie} \\}

\begin{document}
\maketitle
\begin{abstract}
Large Language Models (LLMs) have gathered significant attention due to their impressive performance on a variety of tasks. ChatGPT, developed by OpenAI, is a recent addition to the family of language models and is being called a \textit{disruptive technology} by a few, owing to its human-like text-generation capabilities. Although, many anecdotal examples across the internet have evaluated ChatGPT's strength and weakness, only a few systematic research studies exist. To contribute to the body of literature of systematic research on ChatGPT, we evaluate the performance of ChatGPT on \textit{\textbf{Abstractive Summarization}} by the means of automated metrics and blinded human reviewers. We also build automatic text classifiers to detect ChatGPT generated summaries. \textbf{We found that while text classification algorithms can distinguish between real and generated summaries, humans are unable to distinguish between real summaries and those produced by ChatGPT.}

\end{abstract}





\section{Introduction}
ChatGPT is a recent addition to the family of large language models. Specifically, it is created by extending the work on InstructGPT \cite{ouyang2022training} with a dialog based user-interface that is fine-tuned using Reinforcement Learning with Human Feedback (RLHF) \cite{christiano2017deep} \footnote{https://beta.openai.com/docs/model-index-for-researchers}. ChatGPT is estimated to have reached about $100$ million active users \cite{hu_2023} and is being widely used by business's and customers to accomplish various textual tasks. ChatGPT is easy to use due to it's dialog interface and large scale training via RLHF. Part of its popularity is also abilitieswork, hitherto unseen, such as code generation. ChatGPT has an ability to interact with users in a conversation-style manner and can intelligently "answer follow-up questions, admit its mistakes, challenge incorrect premises and reject inappropriate requests" \cite{schulman2022chatgpt}

However, despite its seemingly magical abilities, anecdotal reports on ChatGPT have suggested significantly remaining challenges - for example it fails in elementary mathematical tasks \citep{frieder2023mathematical}. Open AI has listed many limitations of ChatGPT on its website expressing via a tweet, the CEO tweeted \say{Its a mistake to be relying on ChatGPT for anything important} \cite{altman_2022}. Despite this, there are commercial web based applications that are running on ChatGPT as background. ChatGPT has been found to hallucinate knowledge and frame answers confidently and incorrectly \citep{shen2023chatgpt, thorp2023chatgpt}. Despite its widespread usage systematic studies of ChatGPT's evaluation are scarce. Towards this end, in this study, we evaluate if summaries generated by ChatGPT can be distinguished from real summaries by humans and text-classification algorithms and evaluate generated summaries through automated metrics.  We first prepare a dataset of $50$ summaries generated by ChatGPT and then analyse how ChatGPT summaries are scored across various automated metrics, then evaluate if blinded human reviewers can distinguish between real and generated summaries and finally build a text classifier to detect the two sources.
The sections below discuss related work, data gathering, prompting and results.

\section{Related work}
We briefly survey related work in the areas of \textit{ChatGPT} evaluation and enlist some important studies conducted in \textit{Summarization}.

\paragraph{ChatGPT Evaluation}

Since the release of ChatGPT, there have been many anecdotal evaluation of ChatGPT's abilities and weaknesses posted online but ChatGPT strength and weakness is still a small research area. \citet{Jiao2023IsCA} evaluated ChatGPT's ability on \textit{machine translation} and found that \say{ChatGPT performs competitively
with commercial translation products (e.g.,
Google Translate) on high-resource European
languages but lags behind significantly on low-
resource or distant languages.}. \citet{Kung2022PerformanceOC} evaluated ChatGPT's performance on United States Medical License Examination (USMLE) and found that ChatGPT \say {performed at or near the passing threshold for all three exams without any specialized training or reinforcement}. \citet{Bang2023AMM} is perhaps the most comprehensive evaluation of ChatGPT: the subsection on studies on summarization show that \textit{Interactivity}(recursively generating summaries with modfified instruction) highlight that interactivity improves \textit{ROUGE 1} score. \citet{gao2022comparing}'s work on comparing ChatGPT generated scientific abstracts is similar to our work. In this investigation, authors generated and compared $10$ \textit{Abstracts} in specific journal format and evaluated the generated and original abstracts through an AI output detector, a plagiarism detector and blinded human reviewers. The study found that \say{ChatGPT writes believable scientific abstracts, though with completely generated data. These are original without any plagiarism detected but are often identifiable using an AI output detector and skeptical human reviewers.}

\paragraph{Text Summarization}
Summarization is a task of shortening a large text to a smaller version while retaining the main information.
There are two broad approaches to summarization: \textit{Extractive} and \textit{Abstractive}. Extractive summaries contain \textit{as is} parts from the original text while abstractive summaries can contain novel words and  sentences not found in original text, like human-written summaries. Neural sequence-to-sequence models can generate \emph{abstractive} summaries (meaning the summary generated is not limited to selecting and rearranging text from the original passage) and there is substantial literature on Neural Abstractive Summarization\citep{see2017get, rush2015neural, nallapati2016abstractive, chopra2016abstractive, lewis2019bart}. Recent approaches have seen Large Language Models(LLM) being utilised \citet{devlin2018bert, Yang2019XLNetGA, Raffel2019ExploringTL, Liu2019RoBERTaAR} and performance on summarization reported as standard. In this work, we focus on generating abstractive summaries from ChatGPT and compare it to original summaries. We discuss the precise instructions and dataset curated in section $3$ and $4$ below.

\section{Dataset}
We utilise $50$ \textit{articles} from CNN News/Daily News Dataset \cite{Nallapati2016AbstractiveTS, hermann2015teaching}. The articles were selected from version $3.0.0$ of the test-set. CNN/Daily News Dataset contains $287,226$ training pairs, $13,368$ validation pairs and $11,490$ test pairs. This dataset contains multi-sentence summaries ordered according to the events described in the source article. To prepare a dataset of generated summaries real summaries, we prompted ChatGPT Feb 13 Version \citep{openaiChatGPTRelease} with $50$ \textit{articles} from the CNN News/Daily News Dataset \cite{Nallapati2016AbstractiveTS, hermann2015teaching} with the following prompt: \say{Generate a short summary of the following paragraph in as less words as possible}, the details on why we used this prompt are in the paragraph below. We are also releasing the generates summaries at \footnote{github.com/Mayanksoni20/ChatGPT\_summaries}. 

\paragraph{Prompting}
ChatGPT is an conversation style, instruction-based large language model and the output can depend on the prompt provided. This means two instructions with similar goals can produce different outputs. Keeping this in mind, our goal with selecting a prompt was to produce summaries as similar as possible to the real summaries. Towards this goal, we experimented with a few prompts and finally arrived at the prompt highlighted in bold in Table 1. Other prompts rendered summaries which were longer than real summaries.

\begin{table*}[htb]
\centering
    \begin{tabular}{l}
        \hline
         Prompt Examples \\
         \hline 
         \\
         \textbf{\textit{Generate a short summary of the following paragraph in as less words as possible}} \\
         $\Uparrow$ \\
         \textit{Generate abstractive summary of the following paragraph} \\
         $\Uparrow$ \\
         \textit{Generate highlight for the following paragraph}\\
         $\Uparrow$ \\
         \textit{Reproduce the abstractive summary in as less words as possible} \\
         $\Uparrow$ \\
         \textit{Reproduce the abstractive summary in a maximum of 5 sentences}\\
         $\Uparrow$ \\
         \textit{Generate abstractive summary of the following paragraph in 60 words or less} \\
         $\Uparrow$ \\
         \textit{Produce a short summarize of the following paragraph}\\
         $\Uparrow$ \\
         \textit{Summarize of the following paragraph in 200 words} \\
    \end{tabular}
    \caption{Prompt trials for ChatGPT to obtain responses similar on CNN/Dailymail dataset summaries.}
\end{table*}

\section{Results}
We present the results obtained in automated metrics, human evaluation and text detection in below paragraphs.

\paragraph{Automated Metrics}
To begin with evaluation, we measure ChatGPT's performance on abstractive summarization by comparing original and generated summaries. We utilize the following metrics: ROUGE \cite{lin2004rouge} reporting the F1 scores for ROUGE-1, ROUGE-2 and ROUGE-L and METEOR \cite{banarjee2005}), measuring the word-overlap, bigram-overlap, longest common sequence and unigram overlap between the real and generated summaries. We obtain the scores using HuggingFace evaluate library \footnote{https://github.com/huggingface/evaluate} . We report the results in Table 2. The ROUGE and METEOR scores obtained will be compared to other baseline models in the future versions. 

\begin{table}[ht]
    \centering
    \resizebox{\columnwidth}{!}{%
    \begin{tabular}{ccccc}
     ROUGE-1 & ROUGE-2 & ROUGE-L & ROUGELSUM & METEOR \\ 
     \hline \\
     0.30& 0.11& 0.20& 0.21& 0.35
    \end{tabular}}
    \caption{Automated metrics results between real and generated summaries}
\end{table}

\paragraph{Human Evaluation}
Next, we wanted to evaluate if blinded reviewers were able to distinguish between generated and original summaries. The task of the reviewers was formulated as follows: we asked reviewers to first read the news article followed by reading the summary and then click on either of two radio  buttons guessing if the summary was generated by ChatGPT or human. Figure 1 shows a design of the user interface based on google forms. Two volunteers (native speakers of english) were given $50$ random summaries each. The randomisation worked in such a way that the same volunteer did not receive both (generated and original) summaries of the same article. \textbf{We found that our reviewers were not able to distinguish between generated and human summaries}. The accuracy of human reviewers was reported to be $0.49$. 

\begin{table}[hb]
    \centering
    \resizebox{\columnwidth}{!}{%
    \begin{tabular}{ccc|c}
       &  & \multicolumn{2}{c}{\textit{\textbf{Truth}}}\\
    \hline
         &  & Generated & Original \\
    \hline
     \multirow{2}{*}{\textit{\textbf{Reviewers}} } &  Generated  & $33$ & $29$ \\
       &  Original & $20$ & $15$ \\
    \end{tabular}}
    \caption{Reviewer Scores in Identifying Original and Generated Summaries}
\end{table}

Table 3 shows the confusion matrix of the results obtained. Reviewers also commented that they were sure that they were not able to guess between generated and original summaries, highlighting that the generated and original summaries were similar.

\begin{figure*} [ht]
\centering
    \includegraphics[width=12.5cm,height=12.5cm]{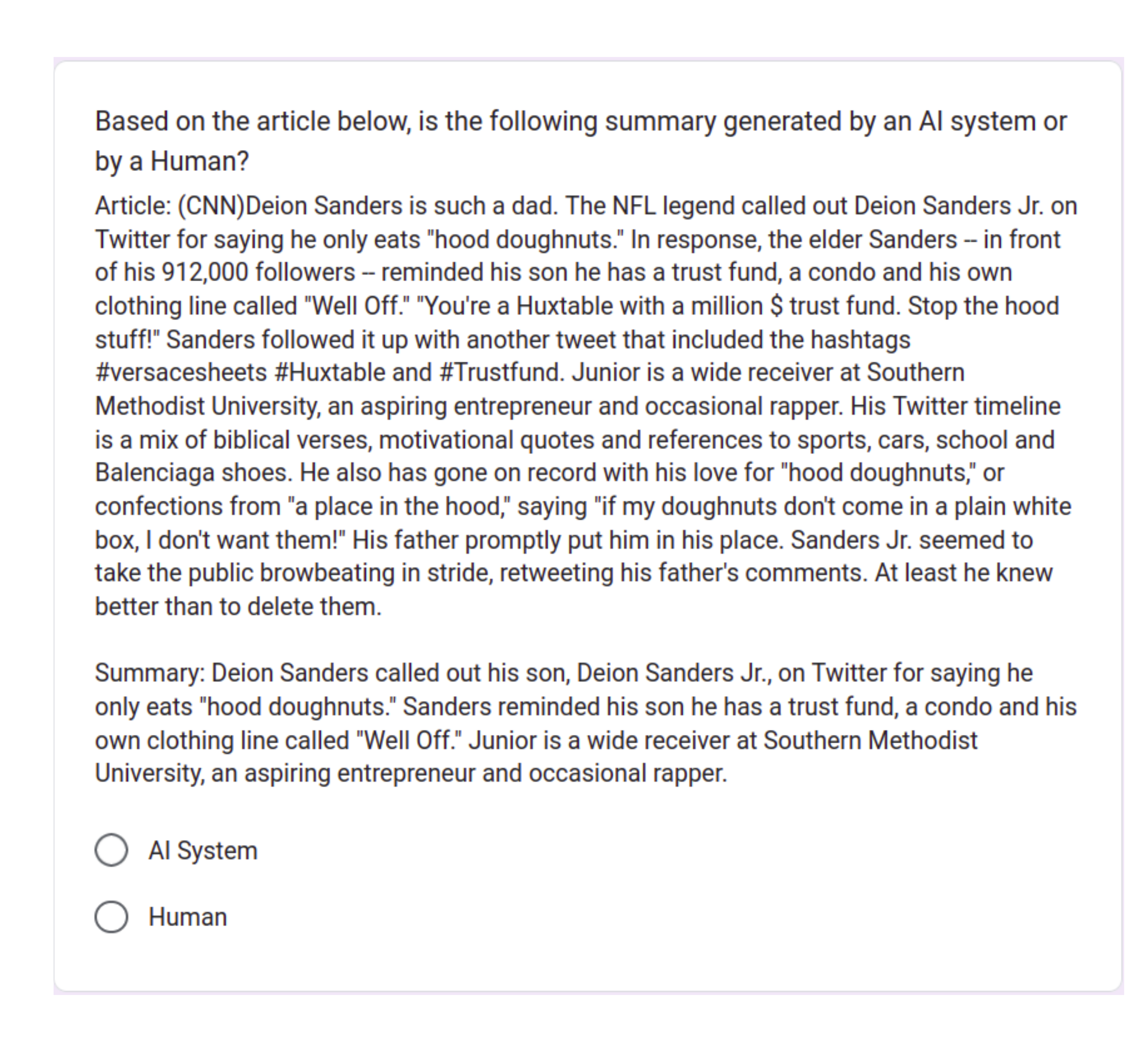}
    \caption{User Interface of Summaries Shown to Human Reviewers} 
\end{figure*}

\paragraph{ChatGPT Summary Detection}
It is very important to be able to distinguish between ChatGPT generated and human produced text to detect ChatGPT generated text. In this next step, we wanted to evaluate if a fine-tuned text classification model is able to distinguish between ChatGPT and Human-generated summaries. We employ DistillBERT from HuggingFace and fine-tune it on our dataset. After Fine-tuning for $2$ epochs with a learning rate of $0.0000002$, we could achieve an accuracy of $90\%$ in distinguishing between generated and original summaries. We also utilise Sentence Embedding \citet{reimers-2019-sentence-bert} and pass them through XGBoost \citet{Chen2016XGBoostAS} and obtain an accuracy of $0.50$. Table 3 shows the results obtained from using the algorithms. We leave the implementation of other algorithms for future work.

\begin{table}[ht!]
    \centering
    \begin{tabular}{c|c|c}
        Algorithm & Accuracy & F-1 Score  \\
        \hline
        SentTrans. +XGB &  0.50 & \textbf{0.60} \\
        Distill-BERT & \textbf{0.90} & 0.33
    \end{tabular}
    \caption{Results of ChatGPT Summary Classification}
\end{table}



\section{Limitations}

The study conducted here has few restrictions.
First, there was a limitation of $50$ on the number of summaries that could be compared. Second, we haven't compared the summaries produced by various prompts, and there may be other prompts that can be used to generate summaries. Third, we have not yet contrasted ChatGPT's summarization performance with that of other models and baselines. Fourth, both of our reviewers were native English speakers; however, it may be worthwhile to examine whether non-native speakers predict real and generated summaries differently. Fifth, the accuracy of automatic summary detection can be increased by using more sophisticated algorithms. 

\section{Discussion and Conclusion}
In this concise study, through comparing 50 real and generated summaries, we found that while text classification algorithms can distinguish between real and generated summaries, humans are unable to distinguish between real summaries and those produced by ChatGPT. Our reviewers were not certain whether a summary was produced by ChatGPT or a human. The reviewers also commented that they were confident that they were incorrectly guessing the source of the summaries. We attribute this to lack of any distinguishing feature between the two sources which in-turn is attributed to our careful selection of prompts to generate summaries as similar as possible to original summaries. Additionally, we were able to identify ChatGPT-generated summaries with a 90\% accuracy. 

\section*{Acknowledgements}
This work was conducted with the financial support of the Science Foundation Ireland Centre for Research Training in Digitally-Enhanced Reality (d-real) under Grant No. 18/CRT/6224 and Science Foundation Ireland ADAPT Centre under Grant. No. 13/RC/2106.

\bibliography{anthology,custom}
\bibliographystyle{acl_natbib}

\appendix



\end{document}